# COST EFFECTIVE APPROACH ON FEATURE SELECTION USING GENETIC ALGORITHMS AND FUZZY LOGIC FOR DIABETES DIAGNOSIS.

E.P.Ephzibah,
School of Information Technology and Engineering, VIT University, Vellore, TamilNadu, India.
ep.ephzibah@vit.ac.in

*ABSTRACT*

*A way to enhance the performance of a model that combines genetic algorithms and fuzzy logic for feature selection and classification is proposed. Early diagnosis of any disease with less cost is preferable. Diabetes is one such disease. Diabetes has become the fourth leading cause of death in developed countries and there is substantial evidence that it is reaching epidemic proportions in many developing and newly industrialized nations. In medical diagnosis, patterns consist of observable symptoms along with the results of diagnostic tests. These tests have various associated costs and risks. In the automated design of pattern classification, the proposed system solves the feature subset selection problem. It is a task of identifying and selecting a useful subset of pattern-representing features from a larger set of features. Using fuzzy rule-based classification system, the proposed system proves to improve the classification accuracy.*

## 1. INTRODUCTION:

Diabetes has become the fourth leading cause of death in developed countries and there is substantial evidence that it is reaching epidemic proportions in many developing and newly industrialized nations, with evidence pointing to avoidable factors such as sedentary lifestyle and poor diet. According to the World Health Organization, it affects around 194 million people worldwide, and that number is expected to increase to at least 300 million by 2025. Prediction of this disease will help to prevent it in its early stage. This work proposes a feature selection model using genetic algorithm, which is efficient to find the best feature subset among the features and the Fuzzy logic rule based classifier, which is used as an effective tool to improve the classification accuracy.

## 2. RELATED WORK:

Feature selection algorithms identify the features that are relevant but not redundant to the solution. The main task is to rank the relevant features based on their fitness values. There are many algorithms that use a greedy search through the solution space. Decision tree algorithms such as Quinlan's ID3 [1] and C4.5 [2], CART proposed in [3], are some of the most successful supervised learning algorithms. Michalski (1980) proposed the AQ learning algorithm. Narendra and Fukunaga (1977) presented a Branch and Bound algorithm. A well known algorithm that relies on relevance evaluation is RELIEF [4]. Subset search algorithms [5] search and capture the goodness of each subset. There are again many algorithms that are exhaustive, heuristic and random search. Clustering algorithms are also used for feature selection process for which ROCK [6], CACTUS [7] are few of them. *Naive Bayes* or Bayes' Rule is the basis for many machine-learning and data mining methods [8]. As for other clinical diagnosis problems, classification systems have been used for disease diagnosis problem. Among many, [9] ToolDiag, RA obtained 50.00% classification accuracy by using IB1- 4 algorithm. [9] WEKA, RA obtained a classification accuracy of 58.50% using InductH algorithm while ToolDiag, RA reached to 60.00% with RBF algorithm. [9] Again, WEKA, RA applied FOIL algorithm to the problem and obtained a classification accuracy of 64.00%. [9] MLP+BP algorithm that was used by ToolDiag, RA reached to 65.60%. [9] The classification accuracies obtained with T2, 1R, IB1c and K* which were applied by WEKA, RA are 68.10%, 71.40%, 74.00% and 76.70%, respectively. [9] Robert Detrano used logistic regression algorithm and obtained 77.0% classification accuracy. Cheung utilized C4.5, Naive Bayes, BNND and BNNF algorithms and reached the classification accuracies 81.11%, 81.48%, 81.11% and 80.96%, respectively [9]. Among the various methods given above the proposed method proves to be more efficient and cost-effective.

## 3. PATTERN RECOGNITION:

A pattern can have a large number of measurable attributes, all of which may not be necessary for uniquely identifying it from other patterns. Good features enhance within-class pattern similarity and between-class pattern dissimilarity. Thus, the selection of measurable attributes is a crucial step in pattern recognition system design. It has been proved that the reason for feature selection is "to curtail the effect of the 'curse of dimensionality' phenomenon on the complexity of the classifier". Feature selection is the process of reducing input data dimension. By reducing dimensionality, feature selection attempts to solve two important problems: facilitate learning (inducing) accurate classifiers, and discover the most "interesting" features, which may provide for better understanding of the problem itself [13].

## 4. NECESSITY FOR COST EFFECTIVE APPROACHES:

People all around the world expect a better approach for the diagnosis of any disease. Feature selection is a technique that reduces or lessens the number of features. In medical world, for any disease to be diagnosed there are some tests to be performed. The disease could be better diagnosed only after getting the results for the tests that are performed. Each and every test can be considered as a feature. To do a particular test certain set of chemicals, equipments are required which can be more expensive. Feature selection informs whether a particular test is necessary for the diagnosis or not. If it is not required that test can be avoided. As a result, when the number of tests gets reduced the cost that is required also gets reduced which helps the common man.

## 5. EVOLUTIONARY COMPUTATION:

The term *evolutionary computation* refers to the study of the foundations and applications of certain heuristic techniques based on the principles of natural evolution. These techniques can be classified into three main categories like genetic algorithms, evolutionary strategies and evolutionary programming. The origin of evolutionary algorithms was an attempt to mimic some of the processes taking place in natural evolution. Although the details of biological evolution are not completely understood (even nowadays), there exist some points supported by strong experimental evidence:

- Evolution is a process operating over chromosomes rather than over organisms. The former are organic tools encoding the structure of a living being, i.e., a creature is "built" decoding a set of chromosomes.
- Natural selection is the mechanism that relates chromosomes with the efficiency of the entity they represent, thus allowing those efficient organisms which are well-adapted to the environment to reproduce more often than those which are not.
- The evolutionary process takes place during the reproduction stage. There exist a large number of reproductive mechanisms in Nature. Most common ones are mutation (that causes the chromosomes of offspring to be different to those of the parents) and recombination (that combines the chromosomes of the parents to produce the offspring).

An Evolutionary Algorithm (EA) is an iterative and stochastic process that operates on a set of individuals (population). Each individual represents a potential solution to the problem being solved. This solution is obtained by means of an encoding/decoding mechanism. Initially, the population is randomly generated (perhaps with the help of a construction heuristic). Every individual in the population is assigned, by means of a fitness function, a measure of its goodness with respect to the problem under consideration. This value is the quantitative information the algorithm uses to guide the search. The whole process is sketched in Figure 3.

*Generate* [P(0)]
$t \leftarrow 0$
**WHILE NOT** *Termination_Criterion* [P($t$)] **DO**

    *Evaluate* [P($t$)]
    P'($t$) $\leftarrow$ *Select* [P($t$)]
    P''($t$) $\leftarrow$ *Apply_Reproduction_Operators* [P'($t$)]
    P($t+1$) $\leftarrow$ *Replace* [P($t$), P''($t$)]
    $t \leftarrow t + 1$

**END**
**RETURN** *Best_Solution*

Figure 1. Skeleton of an Evolutionary Algorithm.

The algorithm comprises three major stages: selection, reproduction and replacement. During the selection stage, a temporary population is created in which the fittest individuals (those corresponding to the best solutions contained in the population) have a higher number of instances than those less fit (natural selection). The reproductive operators are applied to the individuals in this population yielding a new population. Finally, individuals of the original population are substituted by the new created individuals. This replacement usually tries to keep the best individuals deleting the worst ones. The whole process is repeated until a certain termination criterion is achieved (usually after a given number of iterations). EAs are heuristics and thus they do not ensure an optimal solution. The behavior of these algorithms is stochastic so they may potentially present different solutions in different runs of the same algorithm. That's why it is very common to need averaged results when studying some problem and why probabilities of success (or failure), percentages of search extension, etc... are normally used for describing their properties and work.

## 6. SOFT COMPUTING:

Soft computing is a combination of Evolutionary computation, Artificial Neural networks and Fuzzy logic. The guiding principle of soft computing is: Exploit the tolerance for imprecision, uncertainty, and partial truth to achieve tractability, robustness, and low solution cost.

Genetic Algorithms are a family of computational models inspired by evolution. An implementation of a genetic algorithm begins with a population of random chromosomes. The genetic algorithm uses three main types of rules at each step to create the next generation from the current population:

1. *Selection rules* select the individuals, called *parents* that contribute to the population at the next generation.

2. *Crossover rules* combine two parents to form children for the next generation.

3. *Mutation rules* apply random changes to individual parents to form children.

In 1965, L. Zadeh [9] proposed a theory that explains how to formalize "fuzzy" (non-crisp) properties: A crisp property P can be described by a characteristic function $\mu: X \to \{0,1\}$. A fuzzy property can be described as a function $\mu: X \to [0,1]$. The value $\mu(x)$ indicates the degree to which x has the propertyFuzzy logic is used to provide foundations for approximate reasoning using imprecise propositions based on fuzzy set theory, in a way similar to the classical reasoning using precise propositions based on the classical set theory. Fuzzy logic uses the whole interval between 0 (*False*) and 1 (*True*) to describe human reasoning. As a result, fuzzy logic is being applied in rule based automatic controllers.

This work is a model that uses the diabetes dataset and generates the best feature subset using genetic algorithms and fuzzy logic for effective prediction of the disease. The basic structure of the fuzzy inference system (FIS) consists of four components:

- 'fuzzification module' which a determines the membership degrees of the input values in the antecedent fuzzy rules,
- rule base with 'if–then' rules and related membership functions
- 'inference engine' applying algorithms on the rule base and the input data to determine the degree of fulfillment of the output variable, and
- 'defuzzification module' which transforms the fuzzy output into a crisp output value.

For machine control purposes the fuzzy output of these models is defuzzified, or decoded, into one crisp solution, most commonly by calculating the centroïd of area. The working of FIS is as follows. The crisp input is converted in to fuzzy by using fuzzification method. After fuzzification the rule base is formed. The rule base and the database are jointly referred to as the knowledge base. Defuzzification is used to convert fuzzy value to the real world value which is the output. Fuzzy logic control is the main topic of this new field known as Expert Control. Fuzzy logic controls initiated by Mamdani et al [11] are now considered as one of the most important applications of the Fuzzy Set Theory suggested by Zadeh in 1965[12]. It presents the notion of fuzzy set of the ordinary set characterized by a membership function taking the values in the interval [0,1] representing degrees of belonging to the set.[10]

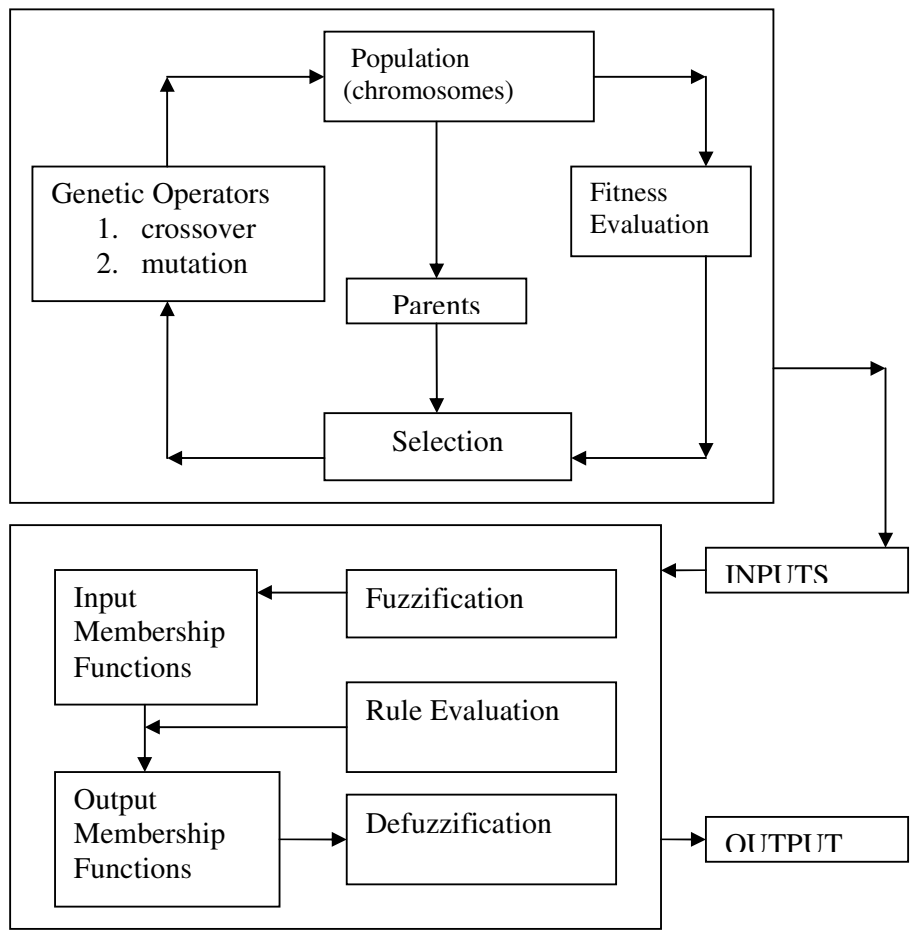

Figure 2.System model combines GA and fuzzy logic

## 7. MATLAB IMPLEMENTATION:

MATLAB essentially supports only one data type, a rectangular matrix of real or complex numeric elements. The main data structures in the GA Toolbox are chromosomes, phenotypes, objective function values and fitness values. The chromosome structure stores an entire population in a single matrix of size *Nind* × *Lind*, where *Nind* is the number of individuals and *Lind* is the length of the chromosome structure. Phenotypes are stored in a matrix of dimensions *Nind* × *Nvar* where *Nvar* is the number of decision variables. An *Nind* × *Nobj* matrix stores the objective function values, where *Nobj* is the number of objectives. Finally, the fitness values are stored in a vector of length *Nind*. In all of these data structures, each row corresponds to a particular individual. The data is taken from the UCI Machine learning Repository where the Pima Indian Diabetes data set is available. This dataset consists of 8 attributes and 769 instances. Out of 8 attributes only three attributes are selected using genetic algorithms. Considering the obtained three attributes and experts knowledge the disease diagnosing process using the MATLAB's Fuzzy logic toolbox for classification is done.

## 8. EXPERIMENTAL RESULTS:

The gatool had been used for GA implementation from MATLAB R2006b. The selection operator is roulette wheel, and the crossover operator is double one-point crossover, and the mutation operator is binary mutation. The size of the population is 40–60. The *Pc* is 0.3–0.9. The *Pm* is 0.01–0.2. The table given below shows that the cost required for the tests is reduced from 46 to 20. It is found to be almost 50% lesser than the original cost.

Table 1. Comparison of cost and accuracy with and without Genetic algorithm approach for Pima Indian Diabetes Dataset.

| Pima Indian Diabetes Dataset | Original set of features | Reduced Feature subset | Accuracy (%) | Cost |
|---|---|---|---|---|
| without GA | 8 | - | 69 | ~=46 |
| with GA | 8 | 3 (±2) | 87 | ~=20 |

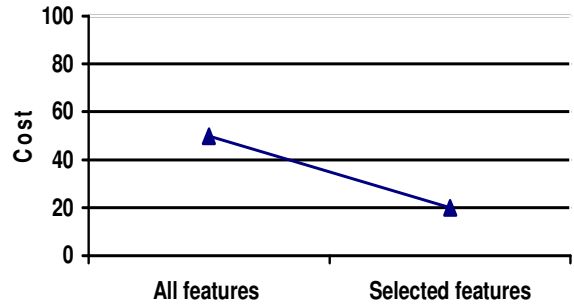

Figure: 3 Cost –Effectiveness graph-Depicting the fall in the cost using the proposed approach.

The inputs for the fuzzy toolbox that consist of the selected features and their membership functions are framed as follows:

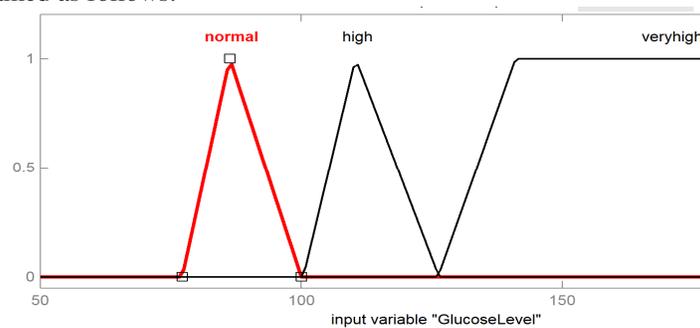

Figure (a)

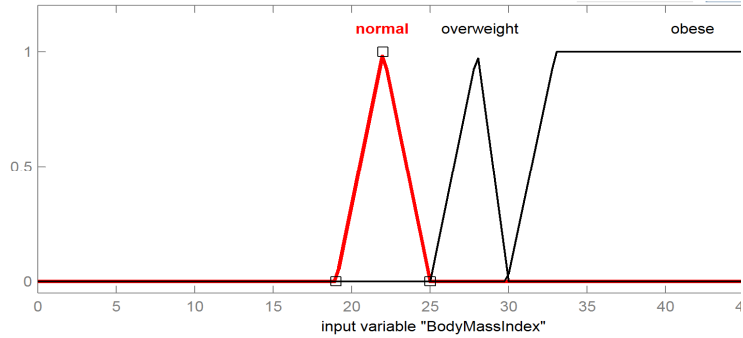
Figure (b)

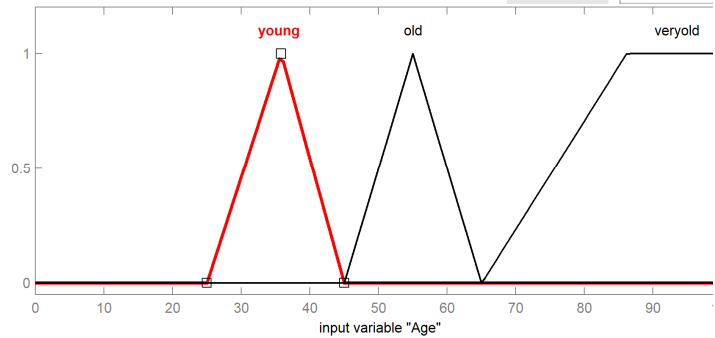
Figure (c )

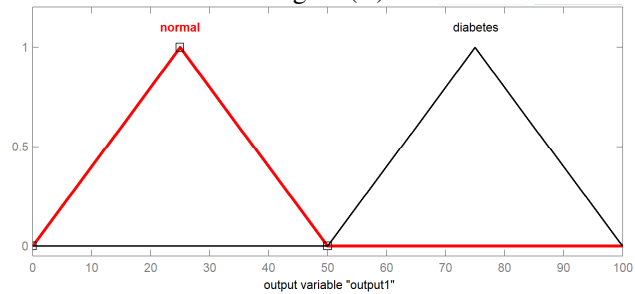
Figure (d)

The figures (a)(b)(c) show the three input variables and their membership functions and the figure (d) is the output that declares whether the patient has diabetes or not.

## 9. CONCLUSION:

This paper has addressed the recently emerged fields of fuzzy and genetic algorithms. The objective of the work is to find the presence of diabetes. The proposed work also helps to minimize the cost and maximize the accuracy. Feature selection or extraction is an important part of pattern recognition and machine learning. With the help of feature selection process, the computation cost decreases and also the classification performance increases. The principle of feature selection has been implemented using the genetic algorithms. It has been shown that applying this principle within a fuzzy logic framework can significantly improve the mechanism's performance to diagnose diabetes in patients. Firstly, a simultaneous mapping is performed based on an appropriateness measure of variables values to each class using suitable membership functions according to each type of feature. Then, simple fuzzy reasoning mechanisms were proposed to deal, in unified way, for classification. For each application task, a validation on real-world problem was performed using Pima Indian Diabetes dataset from the UCI repository. The proposed methodology leads to meaningful results and improves significantly the performance of the system. The experimental results show that this system does quite better than other methods.

**Author**

**E.P.EPHZIBAH** received the M.Sc degree (2001) and M.Phil degree (2004) in Computer science from Madurai Kamaraj University and Bharathidasan University Trichy, India. She is pursuing her Ph.D. programme in the Mother Teresa Women's University, Kodaikanal, India. She is currently working as Assistant Professor in the School of Information Technology and Engineering at VIT University, Vellore, India. Her research interests include DataMining, Feature Selection, Genetic Algorithms, Fuzzy Logic.

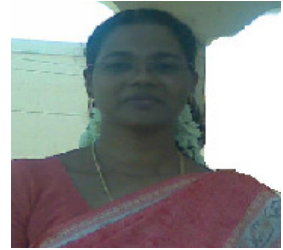